\documentclass{article}

\usepackage[final,main]{neurips_2026}

\usepackage[utf8]{inputenc}
\usepackage[T1]{fontenc}
\usepackage{hyperref}
\usepackage{url}
\usepackage{booktabs}
\usepackage{amsmath}
\usepackage{amssymb}
\usepackage{amsfonts}
\usepackage{amsthm}
\usepackage{graphicx}
\usepackage{algorithm}
\usepackage{algorithmic}
\usepackage[expansion=false]{microtype} 
\usepackage{xcolor}
\usepackage{subcaption}

\newtheorem{theorem}{Theorem}
\newtheorem{proposition}{Proposition}

\title{RoSeViT: Role-Separated Vision Transformers for ARC Visual Reasoning}

\author{%
  Zhiguang Liu \\
  Department of Electrical Engineering\\
  and Computer Science\\
  University of Missouri\\
  Columbia, MO 65211, USA\\
  \texttt{lz7fd@missouri.edu} \\
  \And
  Yi Shang \\
  Department of Electrical Engineering\\
  and Computer Science\\
  University of Missouri\\
  Columbia, MO 65211, USA\\
  \texttt{shangy@missouri.edu} \\
}

\begin{document}

\maketitle

\begin{abstract}
ARC tasks can be formulated as visual transformation problems, where a model
must infer a high-level transformation rule from a small set of demonstrations
and apply it consistently across a grid. While dense Vision Transformers (ViTs)
perform well in this setting, they entangle local visual state and global
transformation logic within a shared token representation, which can hinder
consistent reasoning. In this paper, we introduce RoSeViT, a role-separated
Vision Transformer that explicitly decouples high-level instruction inference
from visual workspace computation. RoSeViT uses controller tokens to aggregate
transformation-level information and workspace tokens to represent the visual
grid, with a structured attention pass in which nonlocal workspace interactions
are routed through the controller. This design is motivated by our formal analysis showing
that enforcing a shared instruction representation can reduce the effective
complexity of visual transformation modeling. RoSeViT outperforms dense ViT baselines under the standard VARC
training and evaluation protocol for ARC. In a controlled backbone replacement setting, it improves ARC-1
accuracy from 54.5 to 56.6, and with recurrent refinement reaches 58.8. When
integrated into the full system, RoSeViT-Ensemble improves performance from 60.4
to 62.6 on ARC-1. These results demonstrate that explicitly separating
instruction and workspace representations leads to more effective visual
reasoning on ARC tasks. GitHub repository: \url{https://github.com/lz7fd/RoSeViT}.
\end{abstract}

\section{Introduction}

ARC tasks ask a system to infer a transformation from a few input-output
demonstrations and apply it to a query grid~\citep{chollet2019measure}. Many
tasks can be viewed as visual transformations governed by a high-level
instruction: copy an object, mirror a shape, extend a pattern, or recolor
matching cells. A successful solver must infer this instruction from
demonstrations and use it consistently across the visual workspace.

VARC shows that ARC can be approached as vision-conditioned grid-to-grid
prediction with test-time adaptation and multi-view
voting~\citep{hu2025arcvisionproblem}. Dense Vision
Transformers~\citep{dosovitskiy2020image} are effective in this setting.
However, they assign the same computational state to all patch tokens: each
patch token may carry local grid state and high-level
transformation instruction simultaneously. This gives dense ViT a larger
instruction-state space than visual transformations require.

Biological and cognitive evidence for separated slow/global and fast/local
processing~\citep{murray2014hierarchy,zeraati2023intrinsic,huntenburg2018large,
bacsar2001gamma,buzsaki2006rhythms,pahor2014theta,tort2009theta}, together with
recent two-channel reasoning systems such as HRM and
TRM~\citep{wang2025hierarchicalreasoningmodel,
jolicoeurmartineau2025morerecursivereasoningtiny}, suggests that the
single-state dense ViT may be suboptimal for visual transformation problems. We
formalize this with the Controller-Route Entropy Reduction Theorem: adding a
controller route decreases the entropy of the available shared-rule solution
route while preserving dense ViT expressivity. The theorem motivates
architectures that keep dense visual feature formation and add an enforced
high-level instruction route.

Motivated by this theorem, we introduce RoSeViT, a role-separated Vision
Transformer for ARC visual reasoning. RoSeViT uses controller tokens to infer
high-level transformation evidence from demonstrations and workspace tokens to
maintain the visual grid state. Each block combines dense visual feature
formation with a structured controller-mediated pass, so nonlocal workspace
communication factors through the controller. Proposition~\ref{prop:low-rank}
shows that this route is globally low-rank.

Under the same VARC training and testing protocol, replacing the dense VARC-ViT
backbone with one-step RoSeViT improves ARC-1 accuracy from 54.5 to 56.6.
Architecture ablations show that the controller bottleneck must be paired with
dense visual feature formation, attention maps are evidence for
controller-mediated routing, and recurrent refinement improves the best single
RoSeViT model to 58.8$\pm$0.7 on ARC-1 and 11.0$\pm$1.0 on ARC-2.
RoSeViT-Ensemble reaches 62.6 on ARC-1 and 13.5 on ARC-2, improving
VARC-Ensemble from 60.4 to 62.6 on ARC-1.

Our contributions are:
\begin{itemize}
    \item \textbf{Controller-Route Entropy Reduction Theorem.} We show that
    adding a controller route reduces the available solution-route entropy for
    shared-rule visual transformations from $\Theta(Ld\log(1/\epsilon))$ to
    $\Theta(d\log(1/\epsilon))$, while RoSeViT contains the dense ViT route;
    whether a trained model uses the route is tested empirically.
    \item \textbf{RoSeViT.} Motivated by this theorem, we introduce a
    role-separated Vision Transformer that realizes the controller route through
    controller instruction tokens and visual workspace tokens. Under the same VARC protocol, RoSeViT improves dense
    VARC-ViT from 54.5 to 56.6 in the controlled backbone comparison, reaches
    58.8 as the best single model, and improves VARC-Ensemble from 60.4 to
    62.6 on ARC-1.
\end{itemize}

\section{Related Work}

\paragraph{ARC and visual reasoning.}
ARC was introduced to measure skill-acquisition and abstraction from a small
number of demonstrations~\citep{chollet2019measure}, with ARC-AGI-2 continuing
this line as a benchmark for frontier reasoning systems~\citep{chollet2025arc}.
Human studies show that ARC remains a demanding visual reasoning benchmark even
for people~\citep{legris2024h}. Recent progress comes from treating ARC as a
vision-conditioned grid-to-grid problem: RE-ARC expands task demonstrations
procedurally~\citep{hodel2024addressingabstractionreasoningcorpus}, and VARC
combines visual augmentation, test-time adaptation, and multi-view
voting~\citep{hu2025arcvisionproblem}. RoSeViT keeps this evaluation stack and
changes the backbone, making the controlled comparison in
Section~\ref{sec:experiments} an architectural test.

\paragraph{Dense ViTs and token roles.}
Transformers~\citep{vaswani2017attention} and Vision
Transformers~\citep{dosovitskiy2020image} use dense self-attention over tokens,
and VARC shows that this design is already strong for ARC visual
reasoning~\citep{hu2025arcvisionproblem}. Dense ViTs can model grids, yet their
patch tokens share the same computational state: every patch token is free to
carry both the inferred transformation and the local workspace state.
Section~\ref{sec:role-separation-theory} formalizes why this creates a larger
instruction-state class than visual transformations require.

\paragraph{Bottlenecks and special tokens.}
Many architectures use special or latent tokens to aggregate information,
including ViT class tokens~\citep{dosovitskiy2020image}, Perceiver-style latent
arrays~\citep{jaegle2021perceiver,jaegle2022perceiverio}, Set Transformer
inducing points~\citep{lee2019set}, Slot Attention
slots~\citep{locatello2020object}, and ViT register
tokens~\citep{darcet2024vision}. These works establish that token roles and
bottlenecks can be useful computational primitives. RoSeViT uses that design
space for a different ARC-specific purpose: the controller bottleneck is
required to infer one high-level transformation instruction and make it
available to the workspace through a structured pass in which nonlocal
workspace communication goes through the controller.

\paragraph{Hierarchical and recurrent reasoning.}
Biological evidence suggests separated slow/global and fast/local processing
timescales~\citep{murray2014hierarchy,zeraati2023intrinsic,huntenburg2018large,
bacsar2001gamma,buzsaki2006rhythms,pahor2014theta,tort2009theta}. HRM and TRM
bring related hierarchical or recurrent ideas to ARC-style
reasoning~\citep{wang2025hierarchicalreasoningmodel,
jolicoeurmartineau2025morerecursivereasoningtiny}. Our contribution is the
computational account missing from these motivations:
the Controller-Route Entropy Reduction Theorem
(Theorem~\ref{thm:role-complexity}) shows that adding a controller route
decreases solution-route entropy while preserving the dense route, and
the collapse condition explains why nominal two-channel systems still need an
enforced bottleneck. Recurrence in RoSeViT is therefore a performance amplifier
applied after the role-separated backbone is defined.

\paragraph{Test-time adaptation.}
Test-time training improves adaptation under limited supervision and
distribution shift~\citep{sun2020test,akyurek2024surprising}. VARC uses this
idea for ARC task adaptation~\citep{hu2025arcvisionproblem}, and RoSeViT follows
the same protocol. The central attribution claim is therefore the same-protocol
backbone comparison in Table~\ref{tab:controlled}; stronger TTT settings and
ensembling are reported as part of the full ARC evaluation stack.

\section{Visual Transformation Complexity}
\label{sec:role-separation-theory}

\begin{figure}[t]
\centering
\includegraphics[width=\linewidth]{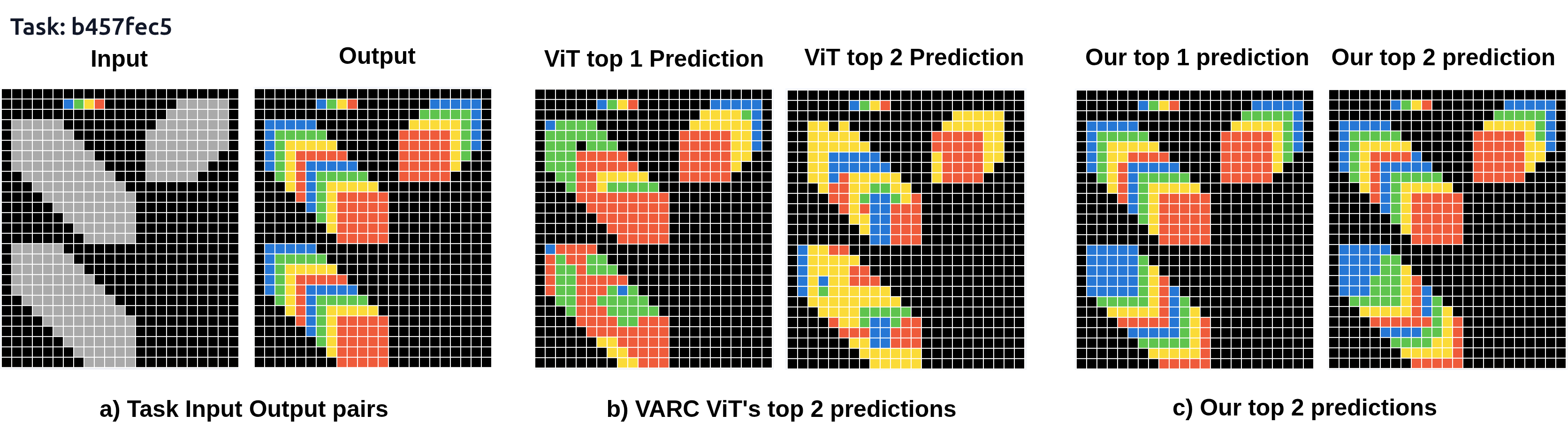}
\caption{ARC visual reasoning is transformation-rule induction, more than visual
completion. VARC-ViT produces locally plausible colored structures, while
RoSeViT c-d-4-TTT3 more consistently applies the inferred transform across repeated
objects.}
\label{fig:intro_example}
\end{figure}

\subsection{Entangled and role-separated routes}

ARC is transformation-rule induction, more than visual completion.
Figure~\ref{fig:intro_example} shows a dense ViT producing locally plausible
colored objects while failing to apply the same transform consistently. In a
dense ViT, each output patch is decoded from a patch token in the same
computational stream, so local visual state and transformation-rule information
are represented together. RoSeViT keeps this dense visual stream and adds
controller tokens as a separated route for transformation-level information.

This design is conservative. The dense route preserves the original VARC-ViT
computation, while the controller route gives shared transformation information
a lower-complexity path through the model. We measure this by solution-route
entropy: the metric entropy of an architecture-provided route that can realize
the shared-rule transformation.

\subsection{Controller-route entropy reduction}

\begin{theorem}[Controller-Route Entropy Reduction Theorem]
\label{thm:role-complexity}
Let visual transformations be governed by one $d$-dimensional shared rule over
$L>1$ workspace sites. Assume the rule-to-output maps are bi-Lipschitz in the
rule variable. A dense single-state patch route represents the rule through
site-wise states
\[
    \mathcal{F}_{\mathrm{site}}
    =
    \{F_{z_1,\ldots,z_L}: z_i\in[0,1]^d\},
\]
with metric entropy
\[
    \log N(\epsilon,\mathcal{F}_{\mathrm{site}},d_\infty)
    =
    \Theta\!\left(Ld\log\frac1\epsilon\right).
\]
RoSeViT preserves this dense route and adds a controller route
\[
    \mathcal{F}_{\mathrm{ctrl}}
    =
    \{F_c: c\in[0,1]^d\},
\]
with metric entropy
\[
    \log N(\epsilon,\mathcal{F}_{\mathrm{ctrl}},d_\infty)
    =
    \Theta\!\left(d\log\frac1\epsilon\right).
\]
For a set of available solution routes $\mathfrak{R}$, define
\[
    \mathcal{H}_\epsilon(\mathfrak{R})
    =
    \min_{\mathcal{F}\in\mathfrak{R}}
    \log N(\epsilon,\mathcal{F},d_\infty).
\]
Then the dense ViT route set
$\mathfrak{R}_{\mathrm{dense}}=\{\mathcal{F}_{\mathrm{site}}\}$ satisfies
\[
    \mathcal{H}_\epsilon(\mathfrak{R}_{\mathrm{dense}})
    =
    \Theta\!\left(Ld\log\frac1\epsilon\right),
\]
whereas the RoSeViT route set
$\mathfrak{R}_{\mathrm{rose}}=
\{\mathcal{F}_{\mathrm{site}},\mathcal{F}_{\mathrm{ctrl}}\}$ satisfies
\[
    \mathcal{H}_\epsilon(\mathfrak{R}_{\mathrm{rose}})
    =
    \Theta\!\left(d\log\frac1\epsilon\right),
\]
and the full RoSeViT class contains the dense ViT class:
\[
    \mathcal{F}_{\mathrm{dense}}\subseteq\mathcal{F}_{\mathrm{rose}}.
\]
Here $\mathcal{F}_{\mathrm{dense}}$ and $\mathcal{F}_{\mathrm{rose}}$ denote
the full dense-ViT and RoSeViT function classes.
Thus adding the controller route decreases the available solution-route entropy
by the workspace-size factor $L$ while preserving dense ViT expressivity.
\end{theorem}

The theorem also explains why a second channel alone is insufficient. The
entropy reduction is obtained only when the architecture enforces the controller
route; if workspace tokens can freely carry independent site-wise rule states,
the effective solution route can collapse back to
$\mathcal{F}_{\mathrm{site}}$. Appendix~\ref{app:theorem-proof} formalizes this
collapse condition.

We use the theorem solely as motivation: it states that a lower-entropy route
exists once the controller route is added, not that optimization selects it,
that the controller encodes the intended rule, or that reduced entropy improves
generalization. Those are empirical claims, tested in
Section~\ref{sec:route-evidence}.
Section~\ref{sec:method} gives the controller-mediated mechanism,
Section~\ref{sec:experiments} tests concrete instantiations, and
Appendix~\ref{app:theorem-proof} provides the proof.

\section{RoSeViT}
\label{sec:method}

\begin{figure}[t]
\centering
\includegraphics[width=0.82\linewidth]{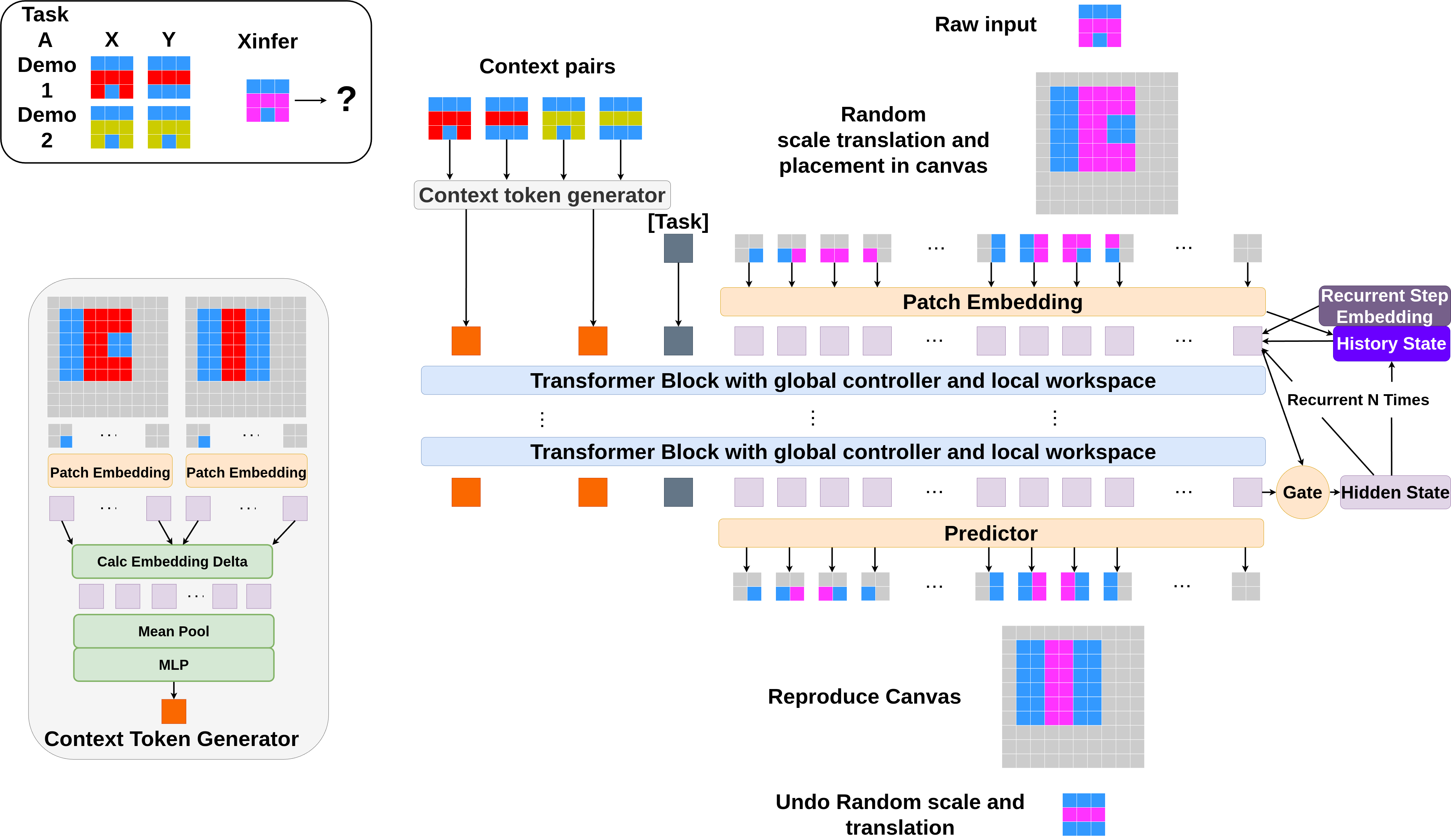}
\caption{Overview of RoSeViT in the VARC visual reasoning pipeline.
Demonstration input-output pairs are transformed into compact controller tokens,
while the query grid is tokenized as a local workspace. RoSeViT blocks
repeatedly combine dense visual feature formation with a structured
controller-mediated pass, and the predicted canvas is mapped back to the ARC
output grid by inverse augmentation.}
\label{fig:overview}
\end{figure}

RoSeViT is designed to instantiate the controller-route entropy reduction in
Theorem~\ref{thm:role-complexity}. The architecture must satisfy three
requirements: preserve the dense visual route, infer a high-level instruction
from demonstrations, and enforce a bottleneck that gives this instruction a
separated route to the workspace. We keep the VARC visual formulation,
augmentation protocol, and output projection, but replace the dense ViT backbone
with a role-separated Transformer block.
Figure~\ref{fig:overview} summarizes the full pipeline.

\subsection{Visual workspace}

An ARC task $T$ provides a small demonstration set
$D^{T}_{\mathrm{demo}}=\{(x_i,y_i)\}_{i=1}^{m}$ and a query input
$x_{\mathrm{infer}}$. Following VARC, each grid is embedded into a larger
augmented canvas using task-preserving transformations such as rotation,
translation, scaling, and fixed borders. Let $\tilde{x}$ denote the augmented
query canvas. We tokenize $\tilde{x}$ using non-overlapping $p\times p$ patches
and map each patch to a $D$-dimensional vector with a linear patch embedding and
a 2D positional embedding. This yields workspace tokens
\[
    w=[w_1,\ldots,w_L]\in\mathbb{R}^{L\times D}.
\]
In our experiments, the canvas is $64\times64$ and $p=2$, so $L=1024$.

\subsection{Controller instruction tokens}

The global controller consists of one task token and one context token per
demonstration pair. The task token $\tau_T$ is learned during offline training
for training tasks; for a new test task, a fresh task token is adapted during
test-time training.

For each demonstration pair $(x_i,y_i)$, we apply the same augmentation family
used for the query grid and embed the augmented input and output grids into
patch tokens $E(\tilde{x}_i),E(\tilde{y}_i)\in\mathbb{R}^{L\times D}$. We form a
pooled transformation descriptor and project it to a context token:
\[
    \bar{\Delta}_i
    =
    \mathrm{Pool}\!\left(E(\tilde{y}_i)-E(\tilde{x}_i)\right),
    \qquad
    c_i=\mathrm{MLP}_{\mathrm{ctx}}(\bar{\Delta}_i).
\]
The controller sequence is
\[
    g=[\tau_T,c_1,\ldots,c_m]\in\mathbb{R}^{P\times D},
    \qquad P=m+1,
\]
and the initial token sequence is
\[
    h^{(0)}=[g;w]\in\mathbb{R}^{(P+L)\times D}.
\]
The controller tokens provide the high-level instruction route: they aggregate
demonstration-level transformation evidence and make it available to the query
workspace. Workspace tokens keep the visual state that will be locally updated
under this instruction.

\subsection{Enforced controller bottleneck}

Each RoSeViT block contains a dense visual pass, a structured role-separated
pass, and an MLP. The dense pass preserves ViT feature formation. The structured
pass then re-imposes role separation: controller queries attend globally, while
workspace queries attend only to the controller tokens and to a small local
workspace neighborhood that includes the token itself,
\[
    \mathcal{A}_{\mathrm{ws}}(i)
    =
    \{1,\ldots,P\}
    \cup
    \{P+j: j\in \mathcal{N}_k(i)\cup\{i\}\},
\]
where $\mathcal{N}_k(i)$ is the 2D local neighborhood of workspace site $i$. In
our main model, $k=8$, corresponding to the eight neighboring patches. Thus,
in the structured pass, high-level instruction information reaches every
workspace token and nonlocal workspace-to-workspace communication passes
through the compact controller.
This is the architectural constraint corresponding to
the collapse condition in Section~\ref{sec:role-separation-theory}: the
attention topology of the structured pass re-imposes role separation through
the controller bottleneck.

\begin{figure}[t]
\centering
\includegraphics[width=0.8\linewidth]{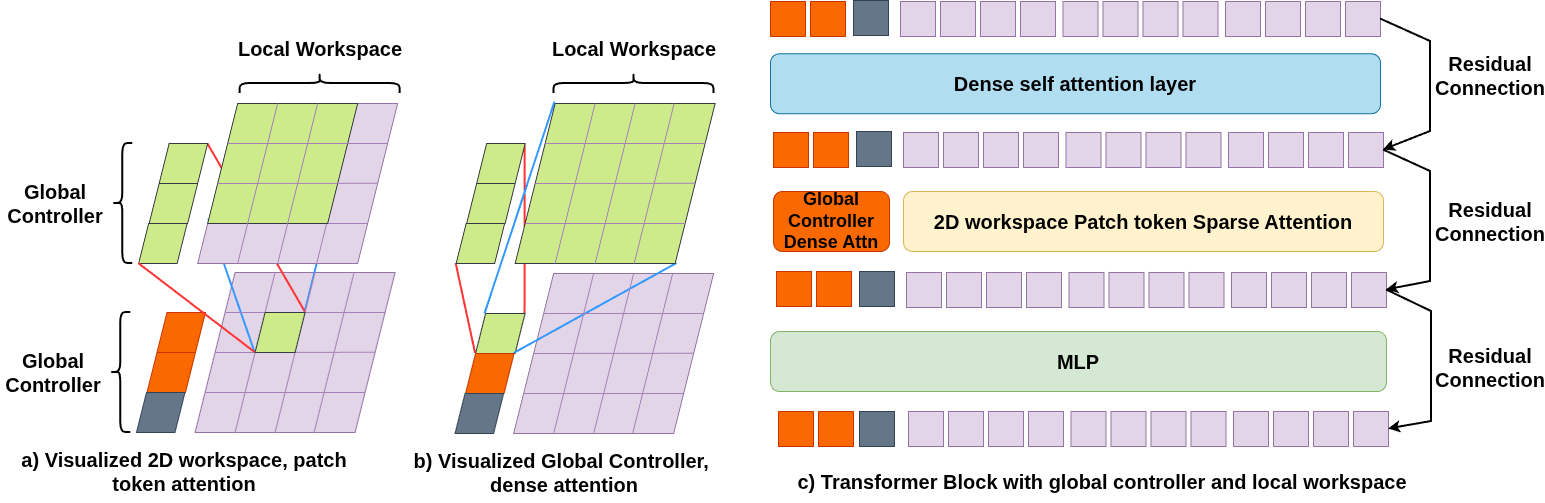}
\caption{Role-separated Transformer block. Controller tokens attend globally,
while workspace tokens attend to the controller and a local spatial
neighborhood. The structured pass repeatedly re-imposes a controller-mediated
bottleneck after dense visual feature mixing.}
\label{fig:block}
\end{figure}

The structured pass gives the added route in Theorem~\ref{thm:role-complexity}:
nonlocal workspace influence factors through the controller, giving the
high-level instruction a designated path through the model. Formally, this route
is globally low-rank and uses
$E_{\mathrm{role}}=P(P+L)+L(P+k+1)$ structured-pass edges, compared with
$E_{\mathrm{dense}}=(P+L)^2$ dense edges; the full statement and proof are in
Appendix~\ref{app:prop-proof}. With $L=1024$, $P\le5$, and $k=8$, each workspace
token attends to at most $P+k+1\le14$ tokens in the structured pass instead of
$P+L=1029$, and the block uses about $1.84\%$ as many structured-pass edges as
dense attention. This edge count describes the route, not our implementation,
which runs the structured pass as dense attention under a mask on top of the
dense pass (measured cost in Section~\ref{sec:experiments}).

\subsection{Recurrent refinement}

We optionally wrap the RoSeViT encoder in a lightweight recurrent refinement
loop to increase effective depth without increasing parameter count. Let
$f_\theta$ denote the stack of role-separated Transformer blocks. At recurrent
step $r$, the model maintains a current token state $h^{(r)}$ and a history
state $s^{(r)}$. It computes
\[
    u^{(r)} = f_\theta(h^{(r)} + s^{(r)}),
    \qquad
    \gamma^{(r)} = \sigma\!\left(W_g[h^{(r)};s^{(r)}]\right),
\]
and applies a gated residual update
\[
    h^{(r+1)} = h^{(r)} + \gamma^{(r)}\odot u^{(r)}.
\]
The history state is updated by an exponential moving average,
\[
    s^{(r+1)} = \alpha s^{(r)} + (1-\alpha)h^{(r+1)}.
\]
This recurrence implements iterative refinement over a fixed ARC query rather
than temporal sequence modeling.

\subsection{Prediction and inverse augmentation}

After the final block or recurrent step, RoSeViT discards the controller tokens
for output decoding and applies a linear head to each workspace token. The head
predicts categorical logits for the corresponding $p\times p$ patch on the
augmented canvas. We assemble the patch logits into a full canvas prediction,
invert the augmentation, crop to the original grid support, and decode the final
ARC output by color-wise argmax.

\section{Experiments}
\label{sec:experiments}

The experiments test the architectural prediction motivated by the
Controller-Route Entropy Reduction Theorem: adding a separated high-level
instruction route while preserving dense visual computation should improve
visual transformation learning. We follow the VARC visual ARC protocol
throughout. Results are pass@2
accuracy on ARC-1 and ARC-2. RoSeViT results are reported as mean over 4 seeds.
Offline training uses ARC-1 training tasks with RE-ARC expansion; ARC-2 is not
used for offline pretraining. At test time, each task is adapted using its
demonstrations and evaluated with 510-view voting. Full protocol, variant, TTT,
and inference details are in Appendix~\ref{app:experiment-details}.

\subsection{Same-protocol test of role separation}
\label{sec:controlled}

Theorem~\ref{thm:role-complexity} motivates the controlled test: replace the
dense VARC-ViT backbone with a one-step RoSeViT d-1 block while keeping the same
training and testing protocol. The input representation, offline data,
augmentation, adaptation budget, 510-view inference, voting rule, and pass@2
metric are unchanged.

Under this controlled setting, RoSeViT d-1 improves ARC-1 accuracy from 54.5 to
56.6, a +2.1 point gain over VARC-ViT. This is the central empirical test of the
high-level-instruction design: with training and testing fixed, the gain
targets the effect of replacing entangled patch-token transformation
representation with controller-mediated instruction routing and workspace
execution.

\textbf{Parameter-matched control.} RoSeViT d-1 has 28M parameters against 18M for VARC-ViT. To separate role
separation from capacity, Table~\ref{tab:controlled} also reports a dense
VARC-ViT widened to 28M under the identical protocol
(Appendix~\ref{app:matched-control}). Widening does not help: the control
reaches 53.6$\pm$1.0 over 8 adaptation seeds (range 52.1--54.9), in line with
the flat-to-declining dense trend 54.5\,$\to$\,53.6\,$\to$\,53.0 at 18M, 28M,
and 66M, and its best seed (54.9) is below the RoSeViT mean. The +2.1 gain over
VARC-ViT and the +3.0 gain over the matched control are therefore
architectural.

\textbf{Compute and latency.} The structured pass is added on top of the dense pass, so RoSeViT is not a
cheaper model. Counted at its sparse edge set (Section~\ref{sec:method}), its
theoretical forward cost is 75.9 GFLOPs, below the 79.4 GFLOPs of the matched
dense control; but no kernel for this access pattern was developed in this work,
our implementation runs the structured pass as dense attention under a mask, and
the actual cost is 97.3 GFLOPs at about $2.1\times$ the latency of the matched
control (Table~\ref{tab:controlled}, Appendix~\ref{app:cost}). The claim of
this paper is role separation, not speed.

\begin{table}[t]
\centering
\caption{Same-training/same-testing backbone comparison (VARC default
adaptation, ARC-1 pass@2) with a parameter-matched dense control and the
measured forward cost (batch 8, H100). Theoretical GFLOPs count the sparse
structured pass; actual GFLOPs are those of our masked dense implementation.}
\label{tab:controlled}
\small
\begin{tabular}{lrrrr}
\toprule
Model & \#param & ARC-1 & GFLOPs theor.\,/\,actual & Latency \\
\midrule
VARC-ViT (dense) & 18M & 54.5 & 53.8\,/\,53.8 & 20.7 ms \\
VARC-ViT, width-matched (dense) & 28M & 53.6 & 79.4\,/\,79.4 & 22.2 ms \\
RoSeViT d-1 (role-separated) & 28M & \textbf{56.6} & 75.9\,/\,97.3 & 46.1 ms \\
\bottomrule
\end{tabular}
\vspace{-6pt}
\end{table}

\subsection{Which architecture realizes the instruction route?}

\begin{figure}[t]
\centering
\includegraphics[width=0.84\linewidth]{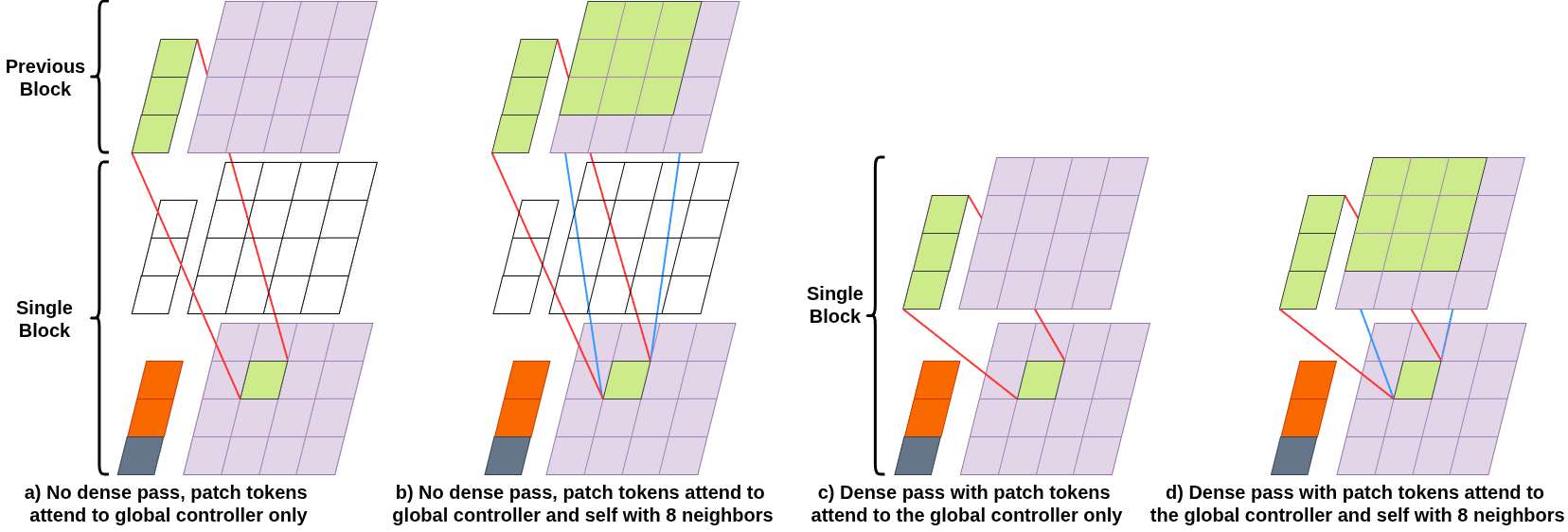}
\caption{RoSeViT block variants used in the architecture ablation. Variants a/b
use only the structured pass, while variants c/d first perform dense visual
feature mixing and then re-impose the controller-mediated bottleneck.}
\label{fig:variants}
\vspace{-6pt}
\end{figure}

The collapse condition says that high/low channels are useful only when the
architecture enforces the instruction route.
Figure~\ref{fig:variants} therefore tests four ways to instantiate the theorem:
a/b use only the
structured controller/workspace pass, while c/d first perform dense visual
feature formation and then re-impose the controller-mediated bottleneck.

The ablation shows why the architecture must realize both sides of the theory.
Pure bottleneck variants a/b enforce separation but are too restrictive for ARC
visual modeling. Variants c/d restore dense visual feature formation and then
re-impose the bottleneck; d adds local workspace neighborhoods and gives the
strongest one-step backbone. Training curves supporting this conclusion are
reported in Appendix~\ref{app:variant-details}. The effective instantiation is
dense visual modeling followed by a structured controller/workspace route.

\subsection{Visible signature of controller-mediated routing}

\begin{figure}[t]
\centering
\includegraphics[width=0.74\linewidth]{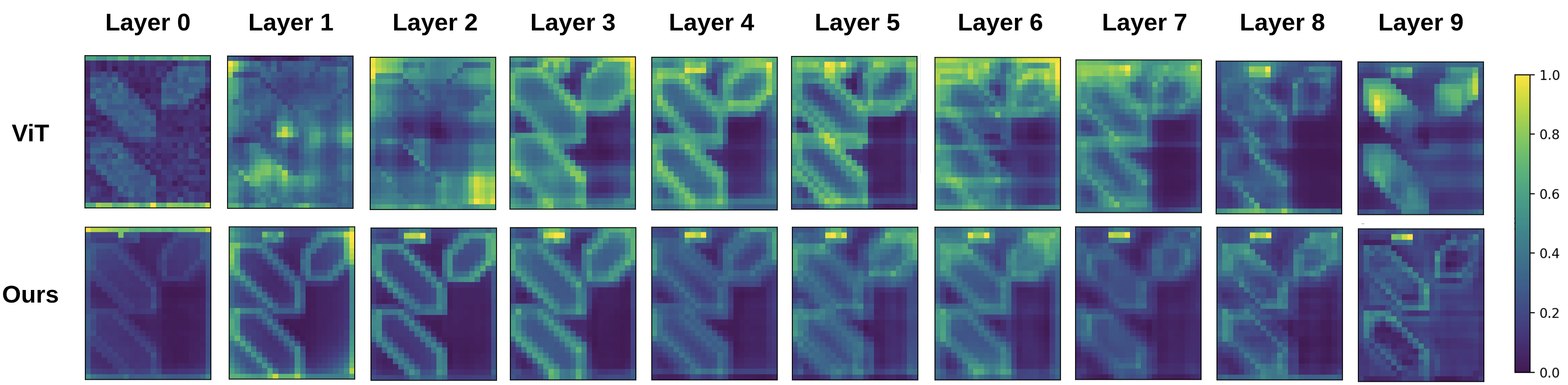}
\caption{Layer-wise attention maps comparing a dense VARC-ViT backbone and
RoSeViT d-1-TTT1 under the same training and testing setting. RoSeViT d-1-TTT1
preserves sharper spatial primitives across depth, matching controller-mediated
high-level instruction routing and local workspace execution.}
\label{fig_6}
\vspace{-6pt}
\end{figure}

Figure~\ref{fig_6} is evidence that the implemented RoSeViT block forms the
controller-mediated route characterized by Proposition~\ref{prop:low-rank}.
Dense ViT attention diffuses through the visual
workspace, consistent with patch tokens mixing local state and transformation
instruction. RoSeViT forms controller-mediated routing: high-level instruction
information is routed through controller tokens, while workspace tokens preserve
sharper spatial primitives for local grid updates.

\subsection{Controller-route evidence: ablation and probe}
\label{sec:route-evidence}

Whether the trained model uses the controller route is an empirical question;
Table~\ref{tab:route} reports two tests on RoSeViT d-1 and the matched dense
control (protocols in Appendix~\ref{app:route-protocols}). First, an
inference-time ablation zeroes the controller embeddings, leaving weights,
masks, dense pass, and local neighborhoods untouched. Removing all conditioning
collapses both models, so collapse itself is not route-specific; the
discriminating results are rule-specificity and non-redundancy. Given the
controllers of a different task, RoSeViT follows the controllers rather than
the input (2.9\% versus 81.3\%), and the task token and the
demonstration-derived context tokens are individually near-essential. After
test-time adaptation, with intact and ablated predictions from the same adapted
weights, the dense control barely needs its conditioning ($-2.5$ pass@2),
whereas RoSeViT still routes rule application through its controllers
($-13.1$). Second, a linear probe (task token excluded) finds the
demonstration-defined rule about twice as decodable from the context tokens
(64.1/53.8/44.0/30.1\% at layers 1/3/6/10) as from the workspace tokens
(33.3/27.7/24.2/16.8\%; chance 0.25\%).

These results establish the controller route as realized, causally necessary,
and rule-specific. They do not establish that the bottleneck prevents rule
information from appearing in workspace tokens: the probe decodes the rule from
workspace tokens well above chance. The measured form of the enforcement claim
is that the controller route is necessary and workspace computation cannot
replace it, not that the workspace carries no rule state.

\begin{table}[t]
\centering
\caption{Controller-route evidence. Top: inference-time ablation, exact
match \% on held-out pairs of training tasks (the dense control has no context
tokens). Middle: ARC-1 pass@2 after adaptation, intact versus ablated
predictions from the same weights (two seeds). Bottom: linear rule probe,
decodability \% of the demonstration-defined rule from context versus
workspace tokens of RoSeViT d-1 by layer (chance 0.25\%).}
\label{tab:route}
\footnotesize
\renewcommand{\arraystretch}{0.88}
\begin{tabular}{lrr}
\toprule
Inference condition (offline, exact match \%) & RoSeViT d-1 & Dense 28M \\
\midrule
intact & 81.3 & 77.9 \\
all conditioning zeroed (task + context / task) & 0.0 & 0.0 \\
wrong-task controllers & 2.9 & -- \\
task token only zeroed & 4.3 & 0.0 \\
context tokens only zeroed & 5.5 & -- \\
\midrule
After TTT (ARC-1 pass@2, 2 seeds) & intact $\to$ ablated & $\Delta$ \\
\midrule
RoSeViT d-1 (controllers zeroed) & 56.3\,/\,56.3 $\to$ 43.3\,/\,43.0 & $-13.1$ \\
Dense 28M (task token zeroed) & 52.9\,/\,54.9 $\to$ 50.3\,/\,52.5 & $-2.5$ \\
\midrule
Rule probe, RoSeViT d-1 (\%; chance 0.25) & context tokens & workspace tokens \\
\midrule
layers 1\,/\,3\,/\,6\,/\,10 & 64.1\,/\,53.8\,/\,44.0\,/\,30.1 & 33.3\,/\,27.7\,/\,24.2\,/\,16.8 \\
\bottomrule
\end{tabular}
\vspace{-6pt}
\end{table}

\subsection{Recurrence and full ARC performance}

After establishing the one-step role-separated backbone, we add recurrence as a
standard performance amplifier for ARC-style reasoning systems, aligned with the
practice in HRM and TRM. Recurrence increases effective computation without
increasing parameter count. The c-2 model applies two recurrent refinement
steps; c-d-4 uses staged training, first forming the controller-only route and
then enabling local workspace neighborhoods.

Table~\ref{tab:ablation} summarizes the main trained variants. The d-1 rows keep
the one-step architecture fixed and stay within a narrow ARC-1 range from 55.5
to 56.6 across adaptation settings. The larger gains appear when
controller-mediated instruction routing is refined recurrently: c-2-TTT2 reaches
58.0 on ARC-1, and
c-d-4-TTT3 reaches 58.8$\pm$0.7 on ARC-1 and 11.0$\pm$1.0 on ARC-2.

\textbf{ARC-2 under the controlled setting.} Under the VARC default adaptation (TTT1), d-1 scores 6.9 on ARC-2 against 8.3
for VARC-ViT. Both backbones are ARC-1 priors (offline training uses ARC-1
tasks with RE-ARC expansion; ARC-2 is never seen): where the prior applies,
role separation extracts more from it, while under the ARC-2 distribution shift
neither prior confers an advantage and performance is set by the adaptation
budget. Accordingly, d-1's ARC-2 score rises with adaptation strength,
6.9\,$\to$\,7.7\,$\to$\,8.3 from TTT1 to TTT3, reaching parity before any
recurrence, and the full system reverses the gap (11.0 versus 8.3). No single
adaptation setting improves both benchmarks for d-1: the setting that reaches
ARC-2 parity (TTT3) costs 1.1 points on ARC-1.

\begin{figure}[t]
\centering
\includegraphics[width=0.5\linewidth]{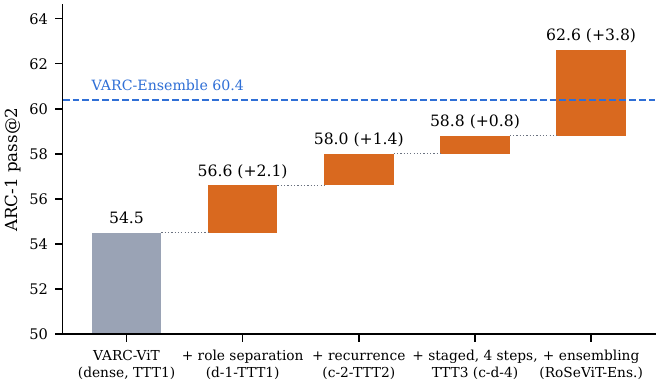}
\caption{ARC-1 gain along the variant ladder of Table~\ref{tab:results};
each step is the difference between consecutive rows, so the recurrence and
staged steps also carry their adaptation-setting change (TTT1\,$\to$\,TTT2
alone moves d-1 by $-0.2$). Dashed line: VARC-Ensemble.}
\label{fig:decomposition}
\vspace{-6pt}
\end{figure}

\textbf{Gain decomposition.} Figure~\ref{fig:decomposition} decomposes the ARC-1 improvement along the
ladder of Table~\ref{tab:results}: role separation alone (the controlled swap)
+2.1, recurrence +1.4, staged training with four recurrent steps and TTT3
+0.8, ensembling +3.8. Voting is part of the fixed protocol in every row, each
amplifier is shared machinery available to both backbones, and the
like-for-like ensemble comparison is 60.4 versus 62.6. The controlled +2.1
remains the central claim.

\begin{table}[t]
\caption{RoSeViT variants and comparison with prior ARC methods (pass@2). All
RoSeViT models use the VARC visual protocol; the controlled architecture
comparison is d-1-TTT1 versus VARC-ViT, and c-d-4-TTT3 is the strongest single
RoSeViT model.}
\label{tab:results}
\begin{subtable}[t]{0.45\linewidth}
\centering
\caption{RoSeViT variants under different recurrent depths and TTT
strategies; the rows form the ladder plotted in Figure~\ref{fig:decomposition}.}
\label{tab:ablation}
\footnotesize
\setlength{\tabcolsep}{4pt}
\begin{tabular}{lrrr}
\toprule
Model & \#param & ARC-1 & ARC-2 \\
\midrule
VARC-ViT       & 18M & 54.5 & 8.3 \\
VARC-ViT (matched) & 28M & 53.6 & -- \\
VARC-ViT-large & 66M & 53.0 & -- \\
\midrule
d-1-TTT1       & 28M & 56.6 & 6.9 \\
d-1-TTT2       & 28M & 56.4 & 7.7 \\
d-1-TTT3       & 28M & 55.5 & 8.3 \\
c-2-TTT2       & 28M & 58.0 & 8.3 \\
c-d-4-TTT3     & 28M & \textbf{58.8} & \textbf{11.0} \\
\bottomrule
\end{tabular}
\end{subtable}\hfill
\begin{subtable}[t]{0.53\linewidth}
\centering
\caption{Comparison to prior ARC methods. RoSeViT improves the single-model
VARC-ViT backbone and remains compatible with VARC-Ensemble.}
\label{tab:comparison}
\footnotesize
\setlength{\tabcolsep}{4pt}
\begin{tabular}{lrrr}
\toprule
Method & \#param & ARC-1 & ARC-2 \\
\midrule
HRM~\citep{wang2025hierarchicalreasoningmodel} & 27M & 40.3 & 5.0 \\
TRM~\citep{jolicoeurmartineau2025morerecursivereasoningtiny} & 7M & 44.6 & 7.8 \\
VARC-ViT~\citep{hu2025arcvisionproblem} & 18M & 54.5 & 8.3 \\
VARC-ViT (matched, ours) & 28M & 53.6 & -- \\
VARC-ViT-large~\citep{hu2025arcvisionproblem} & 66M & 53.0 & -- \\
VARC-Ensemble~\citep{hu2025arcvisionproblem} & 73M & 60.4 & 11.1 \\
\midrule
RoSeViT c-d-4-TTT3 & 28M & \textbf{58.8} & 11.0 \\
RoSeViT-Ensemble & 83M & \textbf{62.6} & \textbf{13.5} \\
\bottomrule
\end{tabular}
\end{subtable}
\vspace{-6pt}
\end{table}

\subsection{Comparison with prior methods}

Table~\ref{tab:comparison} compares RoSeViT with recent ARC solvers and connects
the empirical findings back to the theory. RoSeViT improves over the dense
VARC-ViT backbone, while the larger VARC-ViT underperforms the default dense
model. This supports the thesis that scaling the dense route alone does not add
the separated shared-rule route.
HRM and TRM use
hierarchical or recurrent structure; their lower scores are consistent with, but
not evidence for, the collapse condition, since those systems differ from
RoSeViT in many respects beyond whether a controller-mediated bottleneck is
enforced.

The best single RoSeViT model reaches 58.8$\pm$0.7 on ARC-1 and
11.0$\pm$1.0 on ARC-2, improving over VARC-ViT by +4.3 and +2.7 points.
RoSeViT-Ensemble reaches 62.6 on ARC-1 and 13.5 on ARC-2, improving
VARC-Ensemble from 60.4 to 62.6 on ARC-1. The ensemble result is also
competitive with the reported average human score on ARC-1~\citep{legris2024h};
the central scientific evidence remains the same-protocol comparison, the
architecture ablation, and the controller-route evidence above.

\section{Discussion and Limitations}
\label{sec:discussion}

This paper establishes a specific claim about visual transformation systems:
role separation can preserve dense visual computation while adding a
lower-complexity route for shared transformation information. The
Controller-Route Entropy Reduction Theorem motivates adding this route, while
RoSeViT gives a concrete ViT realization through
controller instruction tokens and visual workspace tokens. The central empirical evidence is the
same-protocol backbone comparison, where RoSeViT d-1 improves over dense
VARC-ViT with the VARC training, adaptation, voting, and evaluation pipeline
held fixed; the full-system results of Table~\ref{tab:comparison} build on it.

The architecture ablation clarifies what kind of role separation is useful.
Pure controller bottlenecks enforce separation but remove too much visual
capacity for ARC grids. The effective design combines dense visual feature
formation with a structured controller-mediated pass that re-imposes the
instruction route. This also sets the claim boundary: recurrence and ensembling
improve the full ARC system, while the central evidence is the effectiveness of
the role-separated Vision Transformer under the same protocol.

The claims of this paper are confined to ARC-type visual transformation
reasoning with a shared per-task rule. The theorem assumes one shared
$d$-dimensional rule; conditional, object-specific, or compositional
transformations exceed this assumption and are untested, as is whether reduced
route entropy correlates quantitatively with generalization. Transfer beyond
ARC is preliminary: an instruction-conditioned maze suite
(Appendix~\ref{app:maze}) shows the same qualitative advantage under a
feedforward, offline protocol, but it is a new suite with two seeds and several
rules unsolved by either model, and we make no claims about embodied or
real-world visual tasks. RoSeViT is not a cheaper model: without a sparse kernel
our implementation runs at about $2.1\times$ the latency of the matched dense
control. RoSeViT also inherits VARC's test-time training and multi-view voting
stack; the controlled comparison isolates the backbone, while final performance
depends on this pipeline. Finally, the controller instruction is latent: the
ablation and probe show the route is necessary and rule-specific, but rule
information remains decodable from workspace tokens, so the bottleneck shapes
rather than fully prevents site-wise rule state.

\section*{Acknowledgements and Disclosure of Funding}
This work received no external funding. Zhiguang Liu was supported by a
graduate research assistantship from the University of Missouri, Columbia; the
compute used for the experiments was paid for by the authors. The authors
declare no competing interests.

\bibliographystyle{plainnat}
\bibliography{rosevit}

@inproceedings{hu2025arcvisionproblem,
      title={{ARC} Is a Vision Problem!}, 
      author={Keya Hu and Ali Cy and Linlu Qiu and Xiaoman Delores Ding and Runqian Wang and Yeyin Eva Zhu and Jacob Andreas and Kaiming He},
      booktitle={Proceedings of the IEEE/CVF Conference on Computer Vision and Pattern Recognition (CVPR)},
      year={2026},
}

@misc{wang2025hierarchicalreasoningmodel,
      title={Hierarchical Reasoning Model}, 
      author={Guan Wang and Jin Li and Yuhao Sun and Xing Chen and Changling Liu and Yue Wu and Meng Lu and Sen Song and Yasin Abbasi Yadkori},
      year={2025},
      eprint={2506.21734},
      archivePrefix={arXiv},
      primaryClass={cs.AI},
      url={https://arxiv.org/abs/2506.21734}, 
}

@misc{jolicoeurmartineau2025morerecursivereasoningtiny,
      title={Less is More: Recursive Reasoning with Tiny Networks}, 
      author={Alexia Jolicoeur-Martineau},
      year={2025},
      eprint={2510.04871},
      archivePrefix={arXiv},
      primaryClass={cs.LG},
      url={https://arxiv.org/abs/2510.04871}, 
}

@misc{hodel2024addressingabstractionreasoningcorpus,
      title={Addressing the Abstraction and Reasoning Corpus via Procedural Example Generation}, 
      author={Michael Hodel},
      year={2024},
      eprint={2404.07353},
      archivePrefix={arXiv},
      primaryClass={cs.LG},
      url={https://arxiv.org/abs/2404.07353}, 
}

@article{murray2014hierarchy,
  title={A hierarchy of intrinsic timescales across primate cortex},
  author={Murray, John D and Bernacchia, Alberto and Freedman, David J and Romo, Ranulfo and Wallis, Jonathan D and Cai, Xinying and Padoa-Schioppa, Camillo and Pasternak, Tatiana and Seo, Hyojung and Lee, Daeyeol and others},
  journal={Nature Neuroscience},
  volume={17},
  number={12},
  pages={1661--1663},
  year={2014},
  publisher={Nature Publishing Group US New York}
}

@article{zeraati2023intrinsic,
  title={Intrinsic timescales in the visual cortex change with selective attention and reflect spatial connectivity},
  author={Zeraati, Roxana and Shi, Yan-Liang and Steinmetz, Nicholas A and Gieselmann, Marc A and Thiele, Alexander and Moore, Tirin and Levina, Anna and Engel, Tatiana A},
  journal={Nature Communications},
  volume={14},
  number={1},
  pages={1858},
  year={2023},
  publisher={Nature Publishing Group UK London}
}

@article{huntenburg2018large,
  title={Large-scale gradients in human cortical organization},
  author={Huntenburg, Julia M and Bazin, Pierre-Louis and Margulies, Daniel S},
  journal={Trends in Cognitive Sciences},
  volume={22},
  number={1},
  pages={21--31},
  year={2018},
  publisher={Elsevier}
}

@article{chollet2019measure,
  title={On the measure of intelligence},
  author={Chollet, Fran{\c{c}}ois},
  journal={arXiv preprint arXiv:1911.01547},
  year={2019}
}

@article{chollet2025arc,
  title={{ARC-AGI-2}: A new challenge for frontier {AI} reasoning systems},
  author={Chollet, Francois and Knoop, Mike and Kamradt, Gregory and Landers, Bryan and Pinkard, Henry},
  journal={arXiv preprint arXiv:2505.11831},
  year={2025}
}

@article{legris2024h,
  title={{H-ARC}: A robust estimate of human performance on the abstraction and reasoning corpus benchmark},
  author={LeGris, Solim and Vong, Wai Keen and Lake, Brenden M and Gureckis, Todd M},
  journal={arXiv preprint arXiv:2409.01374},
  year={2024}
}

@article{vaswani2017attention,
  title={Attention is all you need},
  author={Vaswani, Ashish and Shazeer, Noam and Parmar, Niki and Uszkoreit, Jakob and Jones, Llion and Gomez, Aidan N and Kaiser, {\L}ukasz and Polosukhin, Illia},
  journal={Advances in Neural Information Processing Systems},
  volume={30},
  year={2017}
}

@inproceedings{dosovitskiy2020image,
  title={An image is worth 16x16 words: Transformers for image recognition at scale},
  author={Dosovitskiy, Alexey and Beyer, Lucas and Kolesnikov, Alexander and Weissenborn, Dirk and Zhai, Xiaohua and Unterthiner, Thomas and Dehghani, Mostafa and Minderer, Matthias and Heigold, Georg and Gelly, Sylvain and Uszkoreit, Jakob and Houlsby, Neil},
  booktitle={International Conference on Learning Representations (ICLR)},
  year={2021}
}

@inproceedings{jaegle2021perceiver,
  title={Perceiver: General perception with iterative attention},
  author={Jaegle, Andrew and Gimeno, Felix and Brock, Andrew and Zisserman, Andrew and Vinyals, Oriol and Carreira, Joao},
  booktitle={International Conference on Machine Learning},
  pages={4651--4664},
  year={2021},
  organization={PMLR}
}

@inproceedings{jaegle2022perceiverio,
  title={{Perceiver IO}: A general architecture for structured inputs and outputs},
  author={Jaegle, Andrew and Borgeaud, Sebastian and Alayrac, Jean-Baptiste and Doersch, Carl and Ionescu, Catalin and Ding, David and Koppula, Skanda and Zoran, Daniel and Brock, Andrew and Shelhamer, Evan and Henaff, Olivier and Botvinick, Matthew and Zisserman, Andrew and Vinyals, Oriol and Carreira, Joao},
  booktitle={International Conference on Learning Representations},
  year={2022}
}

@inproceedings{lee2019set,
  title={Set Transformer: A framework for attention-based permutation-invariant neural networks},
  author={Lee, Juho and Lee, Yoonho and Kim, Jungtaek and Kosiorek, Adam and Choi, Seungjin and Teh, Yee Whye},
  booktitle={International Conference on Machine Learning},
  pages={3744--3753},
  year={2019},
  organization={PMLR}
}

@inproceedings{locatello2020object,
  title={Object-centric learning with slot attention},
  author={Locatello, Francesco and Weissenborn, Dirk and Unterthiner, Thomas and Mahendran, Aravindh and Heigold, Georg and Uszkoreit, Jakob and Dosovitskiy, Alexey and Kipf, Thomas},
  booktitle={Advances in Neural Information Processing Systems},
  volume={33},
  pages={11525--11538},
  year={2020}
}

@inproceedings{darcet2024vision,
  title={Vision Transformers Need Registers},
  author={Darcet, Timoth{\'e}e and Oquab, Maxime and Mairal, Julien and Bojanowski, Piotr},
  booktitle={International Conference on Learning Representations},
  year={2024}
}

@inproceedings{lin2017focal,
  title={Focal loss for dense object detection},
  author={Lin, Tsung-Yi and Goyal, Priya and Girshick, Ross and He, Kaiming and Doll{\'a}r, Piotr},
  booktitle={Proceedings of the IEEE International Conference on Computer Vision},
  pages={2980--2988},
  year={2017}
}

@InProceedings{pmlr-v38-lee15a,
  title = 	 {{Deeply-Supervised Nets}},
  author = 	 {Lee, Chen-Yu and Xie, Saining and Gallagher, Patrick and Zhang, Zhengyou and Tu, Zhuowen},
  booktitle = 	 {Proceedings of the Eighteenth International Conference on Artificial Intelligence and Statistics},
  pages = 	 {562--570},
  year = 	 {2015},
  editor = 	 {Lebanon, Guy and Vishwanathan, S. V. N.},
  volume = 	 {38},
  series = 	 {Proceedings of Machine Learning Research},
  address = 	 {San Diego, California, USA},
  month = 	 {09--12 May},
  publisher =    {PMLR},
  url = 	 {https://proceedings.mlr.press/v38/lee15a.html}
}

@article{bacsar2001gamma,
  title={Gamma, alpha, delta, and theta oscillations govern cognitive processes},
  author={Ba{\c{s}}ar, Erol and Ba{\c{s}}ar-Eroglu, Canan and Karaka{\c{s}}, Sirel and Sch{\"u}rmann, Martin},
  journal={International Journal of Psychophysiology},
  volume={39},
  number={2-3},
  pages={241--248},
  year={2001},
  publisher={Elsevier}
}

@book{buzsaki2006rhythms,
  title={Rhythms of the Brain},
  author={Buzs{\'a}ki, Gy{\"o}rgy},
  year={2006},
  publisher={Oxford University Press}
}

@article{pahor2014theta,
  title={Theta--gamma cross-frequency coupling relates to the level of human intelligence},
  author={Pahor, Anja and Jau{\v{s}}ovec, Norbert},
  journal={Intelligence},
  volume={46},
  pages={283--290},
  year={2014},
  publisher={Elsevier}
}

@article{tort2009theta,
  title={Theta--gamma coupling increases during the learning of item--context associations},
  author={Tort, Adriano BL and Komorowski, Robert W and Manns, Joseph R and Kopell, Nancy J and Eichenbaum, Howard},
  journal={Proceedings of the National Academy of Sciences},
  volume={106},
  number={49},
  pages={20942--20947},
  year={2009},
  publisher={National Academy of Sciences}
}

@inproceedings{akyurek2024surprising,
  title={The surprising effectiveness of test-time training for few-shot learning},
  author={Aky{\"u}rek, Ekin and Damani, Mehul and Zweiger, Adam and Qiu, Linlu and Guo, Han and Pari, Jyothish and Kim, Yoon and Andreas, Jacob},
  booktitle={Proceedings of the 42nd International Conference on Machine Learning (ICML)},
  series={Proceedings of Machine Learning Research},
  volume={267},
  pages={942--963},
  year={2025}
}

@inproceedings{sun2020test,
  title={Test-time training with self-supervision for generalization under distribution shifts},
  author={Sun, Yu and Wang, Xiaolong and Liu, Zhuang and Miller, John and Efros, Alexei and Hardt, Moritz},
  booktitle={International Conference on Machine Learning},
  pages={9229--9248},
  year={2020},
  organization={PMLR}
}

\appendix

\section{Proof and Collapse Condition for the Controller-Route Entropy Reduction Theorem}
\label{app:theorem-proof}

We prove the theorem in three steps. First, RoSeViT contains the dense route.
Let a dense ViT block be written as
$D_\theta$ and the corresponding RoSeViT block as
\[
    R_{\theta,\psi}(h)=D_\theta(h)+S_\psi(h),
\]
where $S_\psi$ is the residual controller-mediated route. For any dense ViT
function $f\in\mathcal{F}_{\mathrm{dense}}$, choose the same dense parameters
$\theta$ in RoSeViT and set the output projection of the structured route to
zero, so $S_\psi(h)=0$ for all $h$. The RoSeViT block then computes exactly
$D_\theta$, and repeating this argument for every block gives
\[
    \mathcal{F}_{\mathrm{dense}}\subseteq \mathcal{F}_{\mathrm{rose}} .
\]

Second, we compute the entropy of the two rule routes. Let
$\mathcal{Z}=[0,1]^d$. For a fixed task input, define
\[
    \Phi_i(z)=\phi(x_{\mathcal{N}(i)},p_i,z), \qquad i=1,\ldots,L.
\]
Assume each $\Phi_i$ is bi-Lipschitz in the instruction variable with constants
$0<a\le b$:
\[
    a\|z-z'\|_2
    \le
    \|\Phi_i(z)-\Phi_i(z')\|_2
    \le
    b\|z-z'\|_2 .
\]
Consider two rule routes. A site-wise patch-token route may assign an
independent rule state to each site,
\[
    \mathcal{F}_{\mathrm{site}}
    =
    \left\{
    F_{z_1,\ldots,z_L}: F_i=\Phi_i(z_i),\
    z_i\in\mathcal{Z}
    \right\},
\]
whereas a controller route uses one shared rule variable available to all sites,
\[
    \mathcal{F}_{\mathrm{ctrl}}
    =
    \left\{
    F_z: F_i=\Phi_i(z),\
    z\in\mathcal{Z}
    \right\}.
\]
Under the site-wise metric
\[
    d_\infty(F,G)=\max_i \|F_i-G_i\|_2,
\]
we prove
\[
    \log N(\epsilon,\mathcal{F}_{\mathrm{ctrl}},d_\infty)
    =
    \Theta\!\left(d\log\frac{1}{\epsilon}\right),
    \qquad
    \log N(\epsilon,\mathcal{F}_{\mathrm{site}},d_\infty)
    =
    \Theta\!\left(Ld\log\frac{1}{\epsilon}\right).
\]

The shared-rule family is parameterized by a single instruction variable
$z\in[0,1]^d$.
Since every $\Phi_i$ is $b$-Lipschitz, an $\epsilon/b$-cover of $[0,1]^d$
induces an $\epsilon$-cover of $\mathcal{F}_{\mathrm{ctrl}}$. Conversely, since
each $\Phi_i$ is $a$-Lipschitz from below, an $\epsilon/a$-packing of the unit
cube induces an $\epsilon$-packing of $\mathcal{F}_{\mathrm{ctrl}}$. Standard
covering and packing bounds for the $d$-dimensional cube therefore give
\[
    \log N(\epsilon,\mathcal{F}_{\mathrm{ctrl}},d_\infty)
    =
    \Theta\!\left(d\log\frac{1}{\epsilon}\right).
\]

For the site-wise family, the parameters are
$(z_1,\ldots,z_L)\in[0,1]^{Ld}$. Under $d_\infty$, the induced distance is
equivalent up to constants to $\max_i\|z_i-z_i'\|_2$. Thus a cover is obtained
by independently covering each of the $L$ copies of $[0,1]^d$, and a packing is
obtained by independently packing each copy. This gives
\[
    \log N(\epsilon,\mathcal{F}_{\mathrm{site}},d_\infty)
    =
    \Theta\!\left(Ld\log\frac{1}{\epsilon}\right).
\]

Third, we compute available solution-route entropy. The dense single-state ViT
has the route set
\[
    \mathfrak{R}_{\mathrm{dense}}=\{\mathcal{F}_{\mathrm{site}}\},
\]
so
\[
    \mathcal{H}_\epsilon(\mathfrak{R}_{\mathrm{dense}})
    =
    \log N(\epsilon,\mathcal{F}_{\mathrm{site}},d_\infty)
    =
    \Theta\!\left(Ld\log\frac{1}{\epsilon}\right).
\]
RoSeViT keeps this dense route and adds the controller route:
\[
    \mathfrak{R}_{\mathrm{rose}}
    =
    \{\mathcal{F}_{\mathrm{site}},\mathcal{F}_{\mathrm{ctrl}}\}.
\]
Therefore
\[
    \mathcal{H}_\epsilon(\mathfrak{R}_{\mathrm{rose}})
    =
    \min\{
    \log N(\epsilon,\mathcal{F}_{\mathrm{site}},d_\infty),
    \log N(\epsilon,\mathcal{F}_{\mathrm{ctrl}},d_\infty)
    \}
    =
    \Theta\!\left(d\log\frac{1}{\epsilon}\right).
\]
For $L>1$, adding the controller route decreases the available solution-route
entropy by the workspace-size factor $L$. Together with the dense-route
inclusion above, this proves Theorem~\ref{thm:role-complexity}.

\paragraph{Collapse condition.}
The entropy decrease above comes from adding $\mathcal{F}_{\mathrm{ctrl}}$,
whose free rule variable is shared across all workspace sites. A two-channel
model whose workspace can freely store $z_1,\ldots,z_L$ leaves the site-wise
route available as the effective rule route. The workspace stream can carry the
rule independently at each site, the high-level channel can be bypassed, or both
channels can redundantly represent the same rule information. In all such cases,
the model has no guaranteed solution-route entropy decrease beyond
$\Theta(Ld\log(1/\epsilon))$. An enforced bottleneck is therefore required for a
two-channel design to realize the controller-route entropy reduction.

\section{Controller-Mediated Attention}
\label{app:prop-proof}

\begin{proposition}[Controller-mediated attention is globally low rank]
\label{prop:low-rank}
Consider the structured RoSeViT pass with $P$ controller tokens and $L$
workspace tokens. Ignoring nonlinearities and attention-head constants, write
the controller and workspace updates as
\[
    C' = A_{CC}C + A_{CW}W,
    \qquad
    W' = A_{WC}C' + A^{\mathrm{loc}}_{WW}W .
\]
Then the nonlocal workspace-to-workspace influence factors as
\[
    A_{WC}A_{CW},
\]
and therefore has rank at most $P$. A dense attention layer can realize a
generic workspace-to-workspace influence matrix of rank up to $L$.

If each workspace token attends to $k$ local neighbors in addition to the
controller and itself, the structured pass uses
\[
    E_{\mathrm{role}} = P(P+L) + L(P+k+1)
\]
attention edges, while dense attention over the same tokens uses
\[
    E_{\mathrm{dense}}=(P+L)^2 .
\]
\end{proposition}

\begin{proof}
Substituting the controller update into the workspace update gives
\[
    W' = A_{WC}A_{CC}C + A_{WC}A_{CW}W + A^{\mathrm{loc}}_{WW}W .
\]
The term $A^{\mathrm{loc}}_{WW}W$ contains only local workspace interactions.
The nonlocal workspace-mediated term is $A_{WC}A_{CW}W$. Since
$A_{WC}\in\mathbb{R}^{L\times P}$ and $A_{CW}\in\mathbb{R}^{P\times L}$, the
rank of their product is at most $P$. Dense attention imposes no such
factorization and can represent a generic $L\times L$ interaction matrix.

The edge count follows from the attention mask. Controller tokens attend to all
$P+L$ tokens, giving $P(P+L)$ edges. Each workspace token attends to the $P$
controller tokens, itself, and $k$ local neighbors, giving $L(P+k+1)$ edges.
Dense attention allows all $P+L$ tokens to attend to all $P+L$ tokens, giving
$(P+L)^2$ edges.
\end{proof}

\section{Experimental Details}
\label{app:experiment-details}

\subsection{ARC protocol and data}

We follow the VARC two-stage protocol. Offline training learns one shared model
across ARC-1 training tasks using the task demonstration sets. Following VARC,
each ARC-1 training task is expanded with 1000 RE-ARC demonstrations
\citep{hodel2024addressingabstractionreasoningcorpus}. The inference examples
of ARC-1 training tasks are withheld from offline training and used only for
validation. ARC-2 tasks are not used in offline pretraining.

At test time, each unseen task is adapted from its own demonstration pairs only.
The adapted model is then evaluated on the query inputs. We report pass@2
accuracy: a task is solved when either of the two submitted grids exactly
matches the ground-truth output. Reported RoSeViT accuracies in the main tables
are means over 4 seeds.

\subsection{Shared visual configuration}

All RoSeViT variants use the VARC visual frontend and output projection. Each
grid is embedded into a $64\times64$ augmented canvas, tokenized with
non-overlapping $2\times2$ patches, embedded into dimension 512, and processed
by a depth-10 encoder. The controller contains one task token and one context
token per demonstration pair. During offline training, each training task has a
learned task token. During test-time adaptation, a fresh task token is
initialized for the unseen task and optimized with the model.

\subsection{Architecture variants}
\label{app:variant-details}

We denote each model as \texttt{arch-$r$}, where
\texttt{arch}$\in\{a,b,c,d\}$ is the block design in
Figure~\ref{fig:variants} and $r$ is the number of recurrent refinement steps.
The four block designs are:
\begin{itemize}
    \item \texttt{a}: no dense visual pass; controller tokens attend globally;
    workspace tokens attend only to controller tokens.
    \item \texttt{b}: no dense visual pass; controller tokens attend globally;
    workspace tokens attend to controller tokens and a local $3\times3$
    workspace neighborhood.
    \item \texttt{c}: dense visual self-attention followed by the structured
    pass; in the structured pass, workspace tokens attend only to controller
    tokens.
    \item \texttt{d}: dense visual self-attention followed by the structured
    pass; in the structured pass, workspace tokens attend to controller tokens
    and a local $3\times3$ workspace neighborhood.
\end{itemize}

\begin{figure}[t]
\centering
\includegraphics[width=\linewidth]{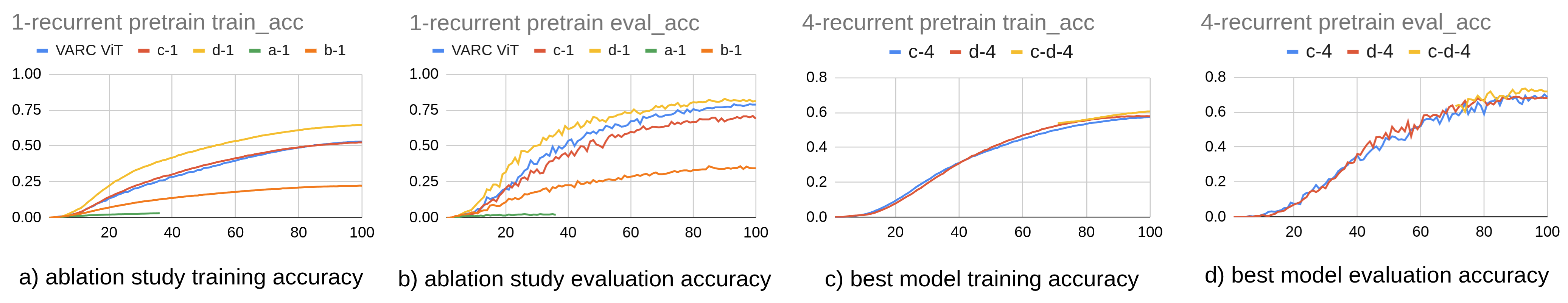}
\caption{Offline training and validation curves for representative RoSeViT
block variants. Variants a/b enforce a strict controller bottleneck from the
start and are too restrictive for ARC visual modeling. Variants c/d keep dense
visual feature formation and then impose the controller/workspace route; d gives
the strongest one-step backbone. For four recurrent steps, staged c-d training
is most stable.}
\label{fig:pretrain_curves}
\end{figure}

Figure~\ref{fig:pretrain_curves} records the architecture search evidence behind
the main-text ablation. Variants a/b show that a bottleneck alone is too
restrictive for learning ARC visual features. Variant c restores dense feature
formation before the structured route, and variant d adds local workspace repair
capacity. For $r=4$, training d end-to-end was less stable in our runs; the best
variant, \texttt{c-d-4}, first trains architecture c for 100 epochs and then
enables the local workspace neighborhood of architecture d for 30 additional
epochs.

\subsection{Offline objectives}

The base offline objective is output-grid cross-entropy after inverse
augmentation and crop to the original grid. Recurrent models may also use a
mid-layer cross-entropy loss for deep supervision \citep{pmlr-v38-lee15a}. In
our implementation, $r\in\{1,2\}$ models are trained without mid-loss, while the
deeper recurrent setting uses mid-loss during offline training.

\subsection{Test-time adaptation}

For each unseen task, test-time training adapts the pretrained model using only
that task's demonstrations, following the TTT paradigm
\citep{akyurek2024surprising,sun2020test}. Each adaptation epoch samples
stochastic task-preserving augmentations of the demonstrations, including the
rotation, translation, scale, and resolution augmentations used by VARC. Unless
otherwise stated, adaptation updates the model end-to-end, including the fresh
task token, and the optimizer state is reset at the start of each task.

We use three adaptation settings:
\begin{itemize}
    \item \texttt{TTT1}: the VARC default adaptation setting, with 100 epochs
    and standard output-grid cross-entropy.
    \item \texttt{TTT2}: a longer, smaller-step adaptation setting with a
    learning rate reduced by $3\times$, 300 epochs, and early termination after
    a 3-epoch streak of 100\% demonstration accuracy.
    \item \texttt{TTT3}: \texttt{TTT2} plus adaptation-only auxiliary objectives
    for exact-grid correction: a grid-level correctness head, pixel-level error
    emphasis for sparse residual mistakes in the spirit of focal loss
    \citep{lin2017focal}, and reward shaping that down-weights output
    cross-entropy and disables mid-loss after an exactly correct demonstration
    prediction is reached.
\end{itemize}

\subsection{Inference and attention visualization}

Inference follows VARC multi-view voting. For each adapted task and query input,
we evaluate 510 augmented views, invert each prediction back to the original
grid coordinates, consolidate predictions by majority vote, and submit the two
most frequent grids as the pass@2 outputs.

Figure~\ref{fig_6} compares RoSeViT d-1-TTT1 and VARC-ViT under the same training
and testing setting. For visualization, the selected ARC example uses ratio-2
upscaling so each original grid cell occupies exactly one patch; this makes the
patch-level attention map align with the pixel grid.

\section{Parameter-Matched Control, Controller-Route Experiments, and Cost Measurement}
\label{app:rebuttal-experiments}

The experiments in this appendix were run with the official VARC codebase and
protocol on an 8$\times$H100 node; adaptation and scoring conventions are those
of Section~\ref{sec:experiments}.

\subsection{Parameter-matched dense control}
\label{app:matched-control}

The width-matched VARC-ViT keeps the default depth and head count and widens the
embedding from 512 to 640 (widths are constrained to multiples of 64 by the
rotary-embedding and kernel divisibility of the codebase); the residual capacity
is placed in the MLP width (\texttt{-{}-mlp-dim} 720). The VARC parameter counter
reports 28.00M, against 28.43M for RoSeViT d-1. Everything else follows the official scripts: 100
offline epochs, VARC default adaptation (TTT1), 510-view voting, pass@2. The
reported 53.6$\pm$1.0 is the mean and standard deviation over 8 adaptation
seeds (range 52.1--54.9); offline training curves are smooth and reach 77.4\%
offline exact match at the final epoch, so the control is not undertrained.

\subsection{Controller ablation}
\label{app:route-protocols}

Controllers of RoSeViT d-1 are one task token and four context tokens. An
ablated inference zeroes the selected controller embeddings at their source
and leaves the weights, attention masks, dense pass, and local workspace
neighborhoods untouched. The offline protocol scores exact match on the
held-out test pairs of the ARC-1 training tasks, the same pairs used for offline
validation, under the training-time evaluation protocol. Conditions:
\emph{all conditioning zeroed} removes the task and context tokens (for the
dense control, which has a task token but no context tokens, it removes the
task token); \emph{batch-mean controllers} replaces each controller with its
mean over the evaluation batch; \emph{wrong-task controllers} swaps in the
controllers of a different training task; \emph{task token only} and
\emph{context tokens only} zero one carrier at a time.

The post-TTT protocol uses the 400 unseen ARC-1 evaluation tasks under the full
adaptation and voting pipeline. For each of two adaptation seeds, the adapted
weights of each task are used twice, once with intact controllers and once with
the controllers (RoSeViT) or the task token (dense control) zeroed, so the
intact and ablated pass@2 values of a seed come from the same weights. The
intact values (56.25/56.25 for RoSeViT d-1, 52.9/54.9 for the dense control)
are consistent with Table~\ref{tab:controlled}.

\subsection{Rule probe}

The probe decodes the demonstration-defined transformation rule (one of the
ARC-1 training rules; chance 0.25\%) from frozen token states of RoSeViT d-1
with a linear classifier, separately from the context tokens and from the
workspace tokens at layers 1, 3, 6, and 10. The task token is excluded from the
probe features to avoid a circular readout of task identity. Because naive
decodability saturates on appearance confounds (the input grid of a training
task identifies the task), the probe uses a dissociation design in which the
rule label must be recovered from states that are not trivially aligned with
appearance; the reported accuracies are on held-out pairs. Rule information is
about twice as decodable from the context tokens as from the workspace tokens
at every depth (64.1/53.8/44.0/30.1 versus 33.3/27.7/24.2/16.8).

\subsection{Compute and latency measurement}
\label{app:cost}

Forward cost was measured on the as-published implementations with a batch of
8 canvases on one H100 GPU: dense VARC-ViT 18M, 53.8 GFLOPs and 20.7 ms; the
width-matched dense 28M control, 79.4 GFLOPs and 22.2 ms; RoSeViT d-1, 97.3
GFLOPs and 46.1 ms. The theoretical figure of 75.9 GFLOPs for RoSeViT d-1
counts the structured pass at the sparse edge set of
Section~\ref{sec:method} (each workspace token attends to at most 14 tokens).
Our implementation does not realize this sparsity: no kernel for this access
pattern was developed in this work, and the structured pass is executed as
dense attention under a boolean mask, which is why actual FLOPs and latency
exceed the matched dense control by about 23\% and $2.1\times$. The difference
between the theoretical and actual cost is an engineering gap, not a property
of the architecture, but the paper makes no efficiency claim.

\section{Preliminary Transfer Evidence: Instruction-Conditioned Maze Suite}
\label{app:maze}

To test whether the architectural advantage appears on a second visual
transformation task family, we built an instruction-conditioned maze suite and
compared parameter-matched RoSeViT and dense ViT backbones under the paper's
protocol without any of the amplifiers: feedforward (no recurrence), offline
training only, no test-time training, no voting, no ensembling, two seeds per
arm. Training data are generated procedurally on the fly, so no result can be
explained by memorization. We report this study as preliminary: the suite is
new, two seeds were run, and four of the eight rules are unsolved by both
models within the training budget.

\begin{figure}[t]
\centering
\includegraphics[width=0.78\linewidth]{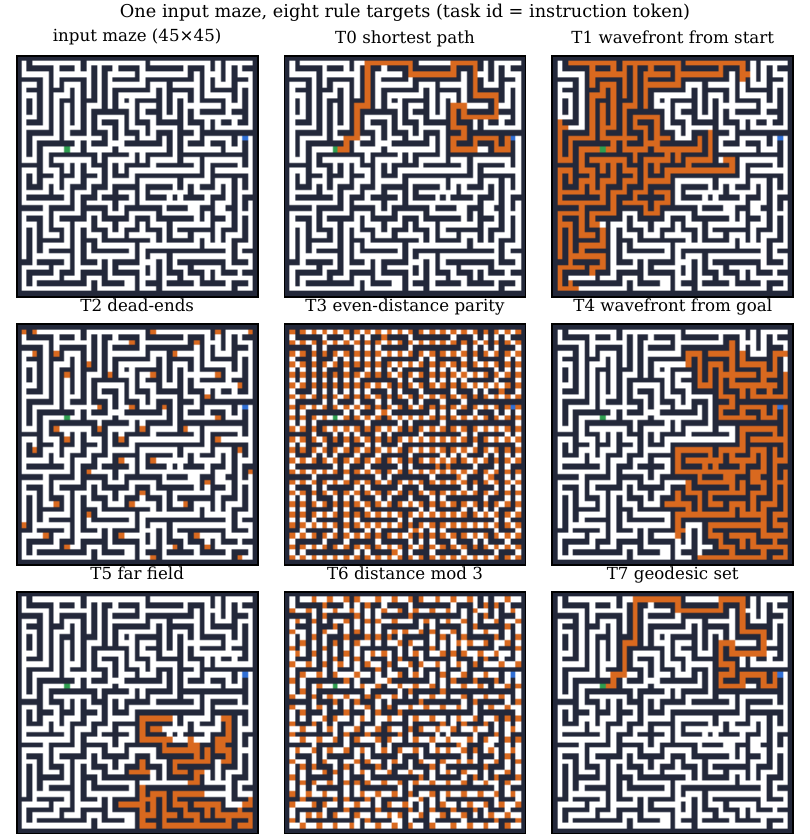}
\caption{The maze suite: one input maze and its eight rule targets. The rule is
given as a task id (an instruction token for RoSeViT; the same embedding as an
extra globally attended token for the dense baseline), so both models must
apply the selected rule to the same input.}
\label{fig:maze-rules}
\end{figure}

\paragraph{Data.}
Each example is a 45$\times$45 corridor maze on a 48$\times$48 canvas
(mirroring the 30-in-32 geometry of HRM Maze-Hard), generated by recursive
backtracking with about 4\% of interior walls knocked out to create loops.
Start and goal are sampled so that the BFS shortest-path length is at least 45
(median about 83). The eight rules are all derived from the BFS distance maps
$d_s$ (from the start) and $d_g$ (from the goal) on the same input
(Figure~\ref{fig:maze-rules}): T0 shortest path (also scored by a strict
validity checker: the predicted cells must form a connected path of optimal
length); T1 wavefront from the start ($d_s\le L/2$); T2 dead-ends (free cells
with exactly one open neighbor, a purely local rule); T3 even-distance parity;
T4 wavefront from the goal ($d_g\le L/2$; the same geometry as T1 with the
opposite anchor, a pure instruction discrimination); T5 far field (top-decile
$d_s$); T6 distance mod 3; T7 geodesic set (all cells on any shortest path).
Scoring is exact match of the full output grid on a fixed held-out set of 128
mazes $\times$ 8 rules.

\paragraph{Models and training.}
Both models are ViTs of width 512 with 8 heads and patch size 2 (a
24$\times$24 grid of 576 workspace tokens). RoSeViT uses depth 8 (33.99M
parameters); each block is a dense pass, a structured pass in which 5
controller tokens attend globally and grid tokens see the controllers and
their 3$\times$3 neighborhood, and an MLP, with zero-initialized gates. The
dense ViT uses standard pre-LN blocks at depth 11 (35.02M parameters), the
depth whose parameter count is nearest to RoSeViT's. The task identity is one learned embedding given
to both models: RoSeViT receives it through its task token, the dense baseline
as an extra token attended globally by all tokens. Both arms use AdamW with
learning rate $3\times10^{-4}$ and cosine decay, bf16, a global batch of 256,
and 40k steps, trained on two H200 GPUs per arm.

\paragraph{Results.}
Table~\ref{tab:maze} reports the full emergence grid, exact match per rule
every 5k steps for every arm and seed. Both models solve the sanity rule T3
immediately. On the local rule T2, RoSeViT is near 0.98 by 10k steps while the
dense baseline stays at 0.00 until about 15k and needs 25--30k steps to master
the same rule. On the global rules T0 and T7, RoSeViT's accuracy emerges at
about 15k steps and climbs to 0.51/0.34 (T0, strict solve) and 0.53/0.38 (T7)
at 40k, whereas the dense baseline reaches at most 0.02 within the same budget
on either seed. T1, T4, T5, and T6 are beyond both models at this budget. The
two architectures therefore climb the same difficulty ladder, with RoSeViT
reaching each rung earlier; at a fixed budget this timing difference appears as
large, sometimes total, performance gaps. For completeness we also evaluated
the public HRM Maze-Hard-1k benchmark: feedforward models of both
architectures score 0\%, matching the HRM paper's own report that
non-recurrent Transformers score 0\% there (shortest paths of length 110 or
more exceed feedforward computational depth), so that benchmark cannot
discriminate between feedforward architectures by design; its 1k-sample
training regime also rewards memorization under recurrent training. We
therefore report the instruction-conditioned suite above as the primary maze
evidence.

\begin{table}[t]
\centering
\caption{Maze suite emergence grid: exact-match accuracy on the held-out set
per rule, arm, and seed, every 5k training steps (feedforward, offline, no
test-time training, no voting). T0 is scored by the strict validity checker.}
\label{tab:maze}
\small
\setlength{\tabcolsep}{4pt}
\begin{tabular}{llrrrrrrrr}
\toprule
Rule & Arm (seed) & 5k & 10k & 15k & 20k & 25k & 30k & 35k & 40k \\
\midrule
T0 shortest path & RoSeViT (42) & .00 & .00 & .02 & .15 & .29 & .40 & .47 & .51 \\
 & RoSeViT (43) & .00 & .00 & .00 & .06 & .24 & .31 & .35 & .34 \\
 & Dense (42) & .00 & .00 & .00 & .00 & .00 & .00 & .02 & .02 \\
 & Dense (43) & .00 & .00 & .00 & .00 & .00 & .00 & .00 & .00 \\
\midrule
T1 wavefront-S & RoSeViT (42) & .00 & .00 & .01 & .01 & .01 & .02 & .02 & .05 \\
 & RoSeViT (43) & .00 & .00 & .00 & .00 & .01 & .00 & .02 & .02 \\
 & Dense (42) & .00 & .00 & .00 & .00 & .00 & .01 & .01 & .01 \\
 & Dense (43) & .00 & .00 & .00 & .00 & .00 & .00 & .00 & .00 \\
\midrule
T2 dead-ends & RoSeViT (42) & .76 & .98 & .99 & 1.00 & 1.00 & 1.00 & 1.00 & 1.00 \\
 & RoSeViT (43) & .82 & .97 & 1.00 & 1.00 & 1.00 & 1.00 & 1.00 & 1.00 \\
 & Dense (42) & .00 & .00 & .35 & .88 & 1.00 & .99 & 1.00 & 1.00 \\
 & Dense (43) & .00 & .00 & .00 & .35 & .91 & .99 & 1.00 & .99 \\
\midrule
T3 parity & RoSeViT (42) & 1.00 & 1.00 & .99 & 1.00 & 1.00 & 1.00 & 1.00 & 1.00 \\
 & RoSeViT (43) & 1.00 & 1.00 & 1.00 & 1.00 & 1.00 & 1.00 & 1.00 & 1.00 \\
 & Dense (42) & 1.00 & .99 & 1.00 & 1.00 & 1.00 & 1.00 & 1.00 & 1.00 \\
 & Dense (43) & .99 & 1.00 & 1.00 & 1.00 & 1.00 & 1.00 & 1.00 & 1.00 \\
\midrule
T4 wavefront-G & RoSeViT (42) & .00 & .00 & .00 & .01 & .00 & .02 & .02 & .02 \\
 & RoSeViT (43) & .00 & .00 & .00 & .00 & .01 & .01 & .01 & .02 \\
 & Dense (42) & .00 & .00 & .00 & .00 & .00 & .00 & .01 & .02 \\
 & Dense (43) & .00 & .00 & .00 & .00 & .00 & .00 & .00 & .00 \\
\midrule
T5 far field & RoSeViT (42) & .00 & .00 & .00 & .01 & .03 & .05 & .02 & .02 \\
 & RoSeViT (43) & .00 & .00 & .00 & .00 & .00 & .03 & .02 & .03 \\
 & Dense (42) & .00 & .00 & .01 & .00 & .00 & .01 & .00 & .01 \\
 & Dense (43) & .00 & .00 & .00 & .00 & .00 & .00 & .00 & .00 \\
\midrule
T6 mod 3 & RoSeViT (42) & .00 & .00 & .00 & .00 & .00 & .00 & .00 & .00 \\
 & RoSeViT (43) & .00 & .00 & .00 & .00 & .00 & .00 & .00 & .00 \\
 & Dense (42) & .00 & .00 & .00 & .00 & .00 & .00 & .00 & .00 \\
 & Dense (43) & .00 & .00 & .00 & .00 & .00 & .00 & .00 & .00 \\
\midrule
T7 geodesic set & RoSeViT (42) & .00 & .00 & .03 & .16 & .30 & .37 & .49 & .53 \\
 & RoSeViT (43) & .00 & .00 & .00 & .08 & .23 & .29 & .37 & .38 \\
 & Dense (42) & .00 & .00 & .00 & .00 & .00 & .00 & .01 & .02 \\
 & Dense (43) & .00 & .00 & .00 & .00 & .00 & .00 & .00 & .00 \\
\bottomrule
\end{tabular}
\end{table}

\clearpage
\section*{NeurIPS Paper Checklist}

\begin{enumerate}

\item {\bf Claims}
    \item[] Question: Do the main claims made in the abstract and introduction accurately reflect the paper's contributions and scope?
    \item[] Answer: \answerYes{}.
    \item[] Justification: The abstract and introduction state the motivating
    theoretical result, the RoSeViT architecture, the controlled backbone
    comparison (with a parameter-matched control), and the full-system ARC
    results; Section~\ref{sec:discussion} states the scope and limitations.

\item {\bf Limitations}
    \item[] Question: Does the paper discuss the limitations of the work performed by the authors?
    \item[] Answer: \answerYes{}.
    \item[] Justification: Section~\ref{sec:discussion} confines the claims to
    ARC-type shared-rule visual transformation reasoning, states the
    single-shared-rule assumption of the theorem, the preliminary status of the
    transfer evidence (Appendix~\ref{app:maze}), the added compute and
    $2.1\times$ latency of our implementation, the dependence on the VARC
    adaptation/voting stack, and that rule information remains decodable from
    workspace tokens.

\item {\bf Theory assumptions and proofs}
    \item[] Question: For each theoretical result, does the paper provide the full set of assumptions and a complete (and correct) proof?
    \item[] Answer: \answerYes{}.
    \item[] Justification: Theorem~\ref{thm:role-complexity} states its
    assumptions, Proposition~\ref{prop:low-rank} states the attention
    factorization setting, and Appendix~\ref{app:theorem-proof} and
    Appendix~\ref{app:prop-proof} provide the proofs.

    \item {\bf Experimental result reproducibility}
    \item[] Question: Does the paper fully disclose all the information needed to reproduce the main experimental results of the paper to the extent that it affects the main claims and/or conclusions of the paper (regardless of whether the code and data are provided or not)?
    \item[] Answer: \answerYes{}.
    \item[] Justification: Section~\ref{sec:experiments} gives the main protocol,
    Appendix~\ref{app:experiment-details} gives training, adaptation, inference,
    and variant details, and the released repository (\url{https://github.com/lz7fd/RoSeViT}) gives
    reproduction commands.

\item {\bf Open access to data and code}
    \item[] Question: Does the paper provide open access to the data and code, with sufficient instructions to faithfully reproduce the main experimental results, as described in supplemental material?
    \item[] Answer: \answerYes{}.
    \item[] Justification: Code, requirements, training scripts, TTT scripts, analysis
    scripts, trained checkpoints, raw prediction archives, and a README with the
    reproduction workflow are released at \url{https://github.com/lz7fd/RoSeViT}.

\item {\bf Experimental setting/details}
    \item[] Question: Does the paper specify all the training and test details (e.g., data splits, hyperparameters, how they were chosen, type of optimizer) necessary to understand the results?
    \item[] Answer: \answerYes{}.
    \item[] Justification: Section~\ref{sec:experiments} specifies the protocol
    needed to interpret the results, and Appendix~\ref{app:experiment-details}
    gives data splits, model variants, offline objectives, TTT settings, and
    inference/voting details.

\item {\bf Experiment statistical significance}
    \item[] Question: Does the paper report error bars suitably and correctly defined or other appropriate information about the statistical significance of the experiments?
    \item[] Answer: \answerYes{}.
    \item[] Justification: RoSeViT results are reported as means over 4 seeds,
    the parameter-matched dense control as mean and standard deviation over 8
    adaptation seeds with its range (Section~\ref{sec:controlled},
    Appendix~\ref{app:matched-control}), the post-TTT
    ablation per seed (Table~\ref{tab:route}), the maze suite per seed
    (Table~\ref{tab:maze}), key standard deviations in the main text for the
    strongest single model, and the released prediction archives allow the reported results to be
    recomputed.

\item {\bf Experiments compute resources}
    \item[] Question: For each experiment, does the paper provide sufficient information on the computer resources (type of compute workers, memory, time of execution) needed to reproduce the experiments?
    \item[] Answer: \answerYes{}.
    \item[] Justification: The reported experiments were run with 8 NVIDIA B200 GPUs;
    the released repository README gives per-script hardware and wall-clock time,
    and the scripts encode 8-GPU training and task-parallel TTT execution. The parameter-matched
    control, the controller ablation and probe, and the cost measurements
    (Appendix~\ref{app:rebuttal-experiments}) were run on an 8$\times$H100 node,
    and the maze suite (Appendix~\ref{app:maze}) on an 8$\times$H200 node with
    two GPUs per arm.
    
\item {\bf Code of ethics}
    \item[] Question: Does the research conducted in the paper conform, in every respect, with the NeurIPS Code of Ethics \url{https://neurips.cc/public/EthicsGuidelines}?
    \item[] Answer: \answerYes{}.
    \item[] Justification: The work uses public ARC-style benchmark data and
    standard model training/evaluation; we are not aware of any deviation from
    the NeurIPS Code of Ethics.

\item {\bf Broader impacts}
    \item[] Question: Does the paper discuss both potential positive societal impacts and negative societal impacts of the work performed?
    \item[] Answer: \answerNA{}.
    \item[] Justification: The work is foundational research on ARC visual
    reasoning and does not introduce a deployed system or application-specific
    societal intervention.
    
\item {\bf Safeguards}
    \item[] Question: Does the paper describe safeguards that have been put in place for responsible release of data or models that have a high risk for misuse (e.g., pre-trained language models, image generators, or scraped datasets)?
    \item[] Answer: \answerNA{}.
    \item[] Justification: The work does not release a high-risk pre-trained
    language model, image generator, scraped dataset, or deployment-facing
    system requiring special safeguards.

\item {\bf Licenses for existing assets}
    \item[] Question: Are the creators or original owners of assets (e.g., code, data, models), used in the paper, properly credited and are the license and terms of use explicitly mentioned and properly respected?
    \item[] Answer: \answerYes{}.
    \item[] Justification: We cite and credit ARC-AGI-1/2, RE-ARC, and VARC.
    ARC-AGI-1/2 are Apache-2.0, RE-ARC is MIT, and the VARC paper/protocol is
    CC BY 4.0.

\item {\bf New assets}
    \item[] Question: Are new assets introduced in the paper well documented and is the documentation provided alongside the assets?
    \item[] Answer: \answerYes{}.
    \item[] Justification: The released repository includes code, scripts, checkpoints,
    raw prediction archives, result visualizations, and a README documenting the
    directory layout, hardware, data layout, and reproduction workflow.

\item {\bf Crowdsourcing and research with human subjects}
    \item[] Question: For crowdsourcing experiments and research with human subjects, does the paper include the full text of instructions given to participants and screenshots, if applicable, as well as details about compensation (if any)? 
    \item[] Answer: \answerNA{}.
    \item[] Justification: This work does not involve crowdsourcing or research
    with human subjects.

\item {\bf Institutional review board (IRB) approvals or equivalent for research with human subjects}
    \item[] Question: Does the paper describe potential risks incurred by study participants, whether such risks were disclosed to the subjects, and whether Institutional Review Board (IRB) approvals (or an equivalent approval/review based on the requirements of your country or institution) were obtained?
    \item[] Answer: \answerNA{}.
    \item[] Justification: This work does not involve human subjects, so IRB or
    equivalent approval is not applicable.

\item {\bf Declaration of LLM usage}
    \item[] Question: Does the paper describe the usage of LLMs if it is an important, original, or non-standard component of the core methods in this research? Note that if the LLM is used only for writing, editing, or formatting purposes and does \emph{not} impact the core methodology, scientific rigor, or originality of the research, declaration is not required.
    \item[] Answer: \answerNA{}.
    \item[] Justification: The core method and experiments do not use LLMs as an
    important, original, or non-standard component.

\end{enumerate}

\end{document}